\documentclass[letterpaper, 10 pt, conference]{ieeeconf}  

\usepackage{amsmath,amsfonts,amssymb}
\usepackage{multirow}
\usepackage{makecell}

\usepackage{graphics} 
\usepackage{graphicx}
\usepackage{soul}
\usepackage{tikz}
\usepackage{xcolor}
\usepackage{wasysym}
\usepackage[utf8]{inputenc}
\usepackage{booktabs}

\usepackage{hyperref}
\usepackage{cleveref}
\usepackage[T1]{fontenc}
\usepackage{caption}

\IEEEoverridecommandlockouts                              

\crefformat{section}{\S#2#1#3} 
\crefformat{subsection}{\S#2#1#3}
\crefformat{subsubsection}{\S#2#1#3}

\title{\LARGE \bf 
 \vspace{-0mm}
Stiffness Copilot: An Impedance Policy for Contact-Rich Teleoperation
\vspace{-2mm}
}

\author{Yeping Wang$^{1,2}$, Zhengtong Xu$^{1,3}$, Pornthep Preechayasomboon$^{1}$, Ben Abbatematteo$^{1}$,\\ Amirhossein H. Memar$^{1}$, Nicholas Colonnese$^{1}$, Sonny Chan$^{1}$
\\[2mm]
\href{http://stiffness-copilot.github.io/}{stiffness-copilot.github.io}
\thanks{$^{1}$All authors are with Meta Reality Labs Research}%
\thanks{$^{2}$Yeping Wang is with University of Wisconsin-Madison, Madison, WI, USA
        {\tt\small yeping@cs.wisc.edu}}%
\thanks{$^{3}$Zhengtong Xu is with Purdue University, West Lafayette, IN, USA}
\thanks{This work was conducted during internships at Meta Reality Labs.}
}

\makeatletter
\let\@oldmaketitle\@maketitle
\renewcommand{\@maketitle}{\@oldmaketitle
   \vspace{3mm}
    \includegraphics[width=7.0in]{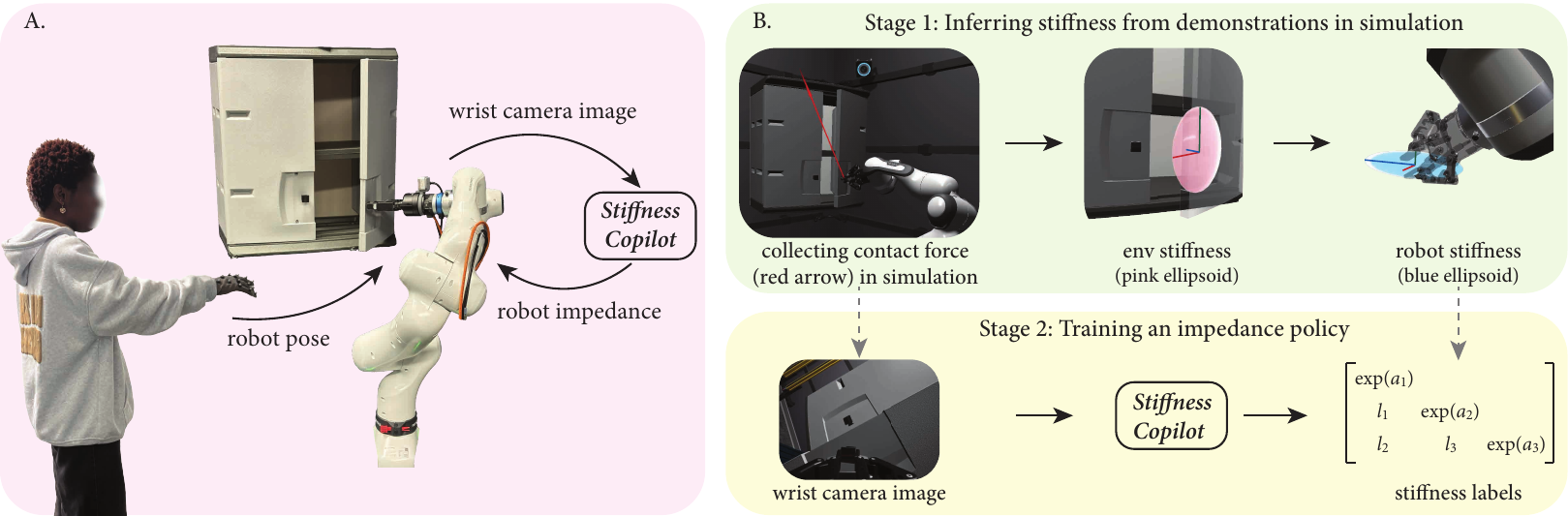}
    \captionof{figure}{We present Stiffness Copilot, a policy that adjusts the robot impedance for contact rich tasks. (A) Stiffness Copilot enables a shared-control teleoperation paradigm in which the human operator commands the robot pose while a vision-based policy adjusts the robot impedance. (B) To train Stiffness Copilot, we collect contact forces in simulation, from which we infer stiffness supervision. The policy is trained on simulated wrist-camera images and deployed zero-shot on real wrist-camera images. }
    \label{fig: teaser}
    \vspace{-4mm}
    }
\newcommand{\algorithmfootnote}[2][\footnotesize]{%
  \let\old@algocf@finish\@algocf@finish
  \def\@algocf@finish{\old@algocf@finish
    \leavevmode\rlap{\begin{minipage}{\linewidth}
    #1#2
    \end{minipage}}%
  }%
}
\makeatother

\begin{document}
 \vspace{-10mm}

\maketitle
\thispagestyle{empty}
\pagestyle{empty}

\addtocounter{figure}{-1} 

\begin{abstract}
In teleoperation of contact-rich manipulation tasks, selecting robot impedance is critical but difficult. The robot must be compliant to avoid damaging the environment, but stiff to remain responsive and to apply force when needed. In this paper, we present Stiffness Copilot, a vision-based policy for shared-control teleoperation in which the operator commands robot pose and the policy adjusts robot impedance online. To train Stiffness Copilot, we first infer direction-dependent stiffness matrices in simulation using privileged contact information. We then use these matrices to supervise a lightweight vision policy that predicts robot stiffness from wrist-camera images and transfers zero-shot to real images at runtime. In a human-subject study, Stiffness Copilot achieved safety comparable to using a constant low stiffness while matching the efficiency of using a constant high stiffness.
\end{abstract}

\vspace{-0mm}
\section{Introduction}
\vspace{-0mm}
Teleoperation is widely used for remote manipulation in unstructured or safety-critical settings and for collecting demonstrations for robot learning. To scale data collection, many systems use lightweight input devices such as VR controllers or a SpaceMouse. These devices are easy to deploy but provide limited or no force feedback, making contact-rich manipulation difficult because human operators cannot sense how the robot loads the environment. In tasks such as door opening, insertion, and wiping, small motion errors can produce large interaction forces, trigger protective stops, and slow operators down.

Variable impedance can reduce this burden. The robot should be compliant enough to tolerate command imprecision and avoid damaging contacts, yet become stiff enough for accurate motion and force transmission. In practice, constant low stiffness improves safety but reduces responsiveness, while constant high stiffness improves efficiency but increases impact risk. This motivates our shared-control approach: the operator commands pose, and a vision-based policy adjusts robot impedance online to match the contact situation. The policy takes wrist-camera images as input because they provide local cues about contact geometry and task phase, which inform the desired stiffness.

In this paper, we present Stiffness Copilot, a shared-control method that automatically adjusts robot stiffness during teleoperation. Appropriate stiffness depends on the contact situation and task phase, so Stiffness Copilot predicts direction-dependent stiffness from wrist-camera images at runtime. We train the policy in two stages. First, in simulation, we use privileged contact and simulator state information to infer stiffness labels for recurring interaction states across demonstrations. Second, we use these labels to supervise a lightweight vision policy, and use a vision foundation model with image augmentation for zero-shot sim-to-real transfer. The resulting policy outputs a symmetric positive definite stiffness matrix and runs in real time.

We conducted a human-subject experiment in which participants teleoperated a robot with constant high stiffness, constant low stiffness, and Stiffness Copilot.
The results indicate that Stiffness Copilot combines the benefits of both baselines. It provides collision safety comparable to low stiffness while matching the task efficiency of high stiffness. It largely avoids large impacts with the environment, yet still transmits sufficient force when needed. Participants also rated Stiffness Copilot as significantly more controllable, trustworthy, predictable, and fluent than both baselines.

The central contribution of this paper is Stiffness Copilot (\cref{sec:method}), a vision-based impedance policy, and the shared-control paradigm it enables. We provide empirical evidence (\cref{sec:evalution}) that this paradigm improves task performance and user experience during teleoperation of contact-rich tasks.

\vspace{-0mm}
\section{Related Works}
\vspace{-0mm}

Stiffness Copilot is a learning-based variable impedance policy for teleoperation. This section reviews three lines of work that inform our approach.

\vspace{-0mm}
\subsection{Learning-Based Variable Impedance Policy}
\vspace{-0mm}

Contact-rich manipulation, such as surface wiping and peg-in-hole assembly, requires regulating interaction dynamics in addition to tracking motion, motivating impedance modulation during execution. Accordingly, many imitation learning methods augment pose prediction with an explicit compliance output.
Works such as \cite{aburub2024learning, hou2025adaptive, kamijo2024learning} train policies to output robot stiffness, regulating interaction via impedance control. In contrast, approaches like \cite{kormushev2011imitation, zhou2025admittance, wu2024tacdiffusion} predict desired forces, often using admittance or hybrid force-position control. Reinforcement learning methods \cite{martin2019variable} have also adopted variable impedance control to autonomously discover compliance strategies for dynamic interaction tasks. However, these autonomous methods cannot be directly used for shared-control teleoperation because the learned stiffness is optimized for the policy's own motion. Applying this trajectory-specific impedance to a human's variable inputs can lead to suboptimal compliance or instability.

To generate stiffness labels in imitation learning, prior work primarily relies on two data collection modalities. The first is kinesthetic teaching \cite{aburub2024learning, hou2025adaptive}, where the operator physically guides the robot. This requires the robot to remain low-stiffness to accept human guidance. As the robot's compliance damps interaction forces, stiffness estimated from such softened signals can be conservative for rigid interactions \cite{hogan2018impedance}. The second approach is teleoperation with manual modulation \cite{kamijo2024learning}, where operators adjust impedance by pressing a button. This restricts labels to predefined templates and makes it difficult to capture continuous full-matrix stiffness profiles.

In contrast, our approach leverages privileged information from simulation to construct stiffness supervision and distill it to a policy. Unlike real-world training, which restricts exploration to compliant regimes for safety, simulation permits safe collection of high-stiffness interaction data. We use explicit object state and contact forces to construct structured stiffness labels from interaction dynamics. This transforms the problem into supervised regression over impedance targets, allowing the policy to learn coupled stiffness behaviors that are difficult for humans to specify manually.

\vspace{-0mm}
\subsection{Variable Impedance in Teleoperation}
\vspace{-0mm}

Determining appropriate impedance parameters for teleoperation remains a challenge. Variable impedance control can be broadly categorized by how stiffness is specified: explicitly by the human operator, implicitly through physiological signals, or automatically through task-based rules.

\textit{Explicit Specification} -- Prior work has developed interfaces that allow users to shape the stiffness ellipsoid using a tablet interface \cite{peternel2021independently}, a foot pedal \cite{klevering2022foot}, or interactive visualizations in virtual reality \cite{rosales2024interactive}. However, manually tuning a full stiffness matrix is cognitively demanding and often requires domain expertise. Recently, Hernandez \textit{et al.} \cite{hernandez2025stiffness} introduced a haptic co-pilot that guides the operator to switch between discrete stiffness levels.

\textit{Implicit Estimation} -- To achieve more natural modulation without manual input, the Tele-Impedance framework \cite{ajoudani2012tele} estimates the human operator's arm stiffness using electromyography and maps it to the robot. This allows the robot to mimic the human's variable stiffness behavior implicitly, though it requires additional physiological sensing hardware.

\textit{Rule-Based Automation} -- Alternatively, semi-autonomous methods reduce operator burden by regulating impedance using sensor feedback and task-specific heuristics. These methods typically adjust stiffness magnitude rather than orientation. A common pattern is to modulate only the stiffness magnitude, for example by increasing stiffness with measured external force to stabilize initial contact \cite{muratore2018enhanced}, adjusting stiffness from shear force to prevent object slip during grasping \cite{popken2023adaptive}, or switching between a small number of impedance modes based on visually detected grasp phases \cite{huang2021semi}. While effective in their target settings, these rules are closely tied to specific sensing cues and hand-crafted phase definitions, which limits their flexibility.

\textit{Learning-Based Estimation} -- Unlike rule-based methods that rely on predefined heuristics for specific tasks, data-driven approaches infer appropriate impedance profiles from human demonstrations. Michel et al. \cite{michel2021bilateral, michel2023learning, michel2024passivity} modulate stiffness inversely proportional to trajectory variance in demonstrations and train Gaussian mixture models to infer stiffness based on sensed forces. However, this method is designed for tasks with continuous interaction with the environment, such as cutting. It also implicitly assumes that demonstrations use one strategy because it infers stiffness from trajectory variance. It remains unclear how to extend this idea to intermittent-contact tasks or tasks with multiple strategies.

\vspace{-0mm}
\subsection{Learning-Based Assistance in Teleoperation}
\vspace{-0mm}

Shared control in teleoperation \cite{li2023classification} is a collaborative control paradigm in which human commands and an assistive policy are blended in real time to drive the remote robot. Recent learning-based approaches first learn such a policy from data and then use it to assist teleoperation. For humanoid multi-contact manipulation, Rouxel et al. \cite{rouxel2024flow} learn a flow-matching policy from teleoperated demonstrations and deploy it in a shared-autonomy mode to provide automatic contact placement and stabilize the robot. Yoneda et al. \cite{yoneda2023noise} and Sun et al. \cite{sun2025flashback} train generative models that infer user intent and combine user inputs with expert-like actions for shared autonomy. In parallel, joint learning frameworks such as Human-Agent Joint Learning \cite{luo2024human, hong2026scoop} and Incrementally Learned Shared Autonomy \cite{tao2024incremental} close the loop between data collection and assistance. They start from weak or synthetic assistive policies, use them to support teleoperation and collect higher quality demonstrations, and gradually adapt the shared-control policy toward higher levels of autonomy while keeping the human in the loop.

\vspace{-0mm}
\section{Stiffness Copilot} \label{sec:method}
\vspace{-0mm}

In this section, we formalize the problem and describe the training and deployment of Stiffness Copilot.

\vspace{-0mm}
\subsection{Problem Formulation}
\vspace{-0mm}

During teleoperation, the operator commands the end-effector pose and Stiffness Copilot modulates impedance. We learn the translational stiffness matrix of a task-space impedance controller and keep rotational stiffness and damping fixed. The predicted matrix specifies how the end effector yields to or transmits contact forces across directions.

Stiffness Copilot predicts a normalized translational stiffness profile $\mathbf{\tilde{K}} \in \mathbb{R}^{3\times 3}$, a symmetric positive definite matrix with eigenvalues in $[0,1]$. At deployment, we obtain the stiffness matrix $\mathbf{K} \in \mathbb{R}^{3\times 3}$ by linearly mapping these eigenvalues to $[k_{\min}, k_{\max}]$ in N/m and reconstructing a symmetric positive definite matrix with the same eigenvectors.

We predict a normalized stiffness profile $\mathbf{\tilde{K}}$ rather than a stiffness matrix in absolute units. In human-in-the-loop teleoperation, there is no single ``correct'' stiffness matrix: different choices trade off tracking accuracy, responsiveness, safety, and operator comfort. Normalization lets the policy learn directional stiffness allocation and relative stiffness across directions, while $k_{\min}$ and $k_{\max}$ set the robot-specific scale at deployment.

\vspace{-0mm}
\subsection{Demonstration in Simulation}
\vspace{-0mm}
We collect demonstrations in a simulation-based teleoperation environment with randomized object placements. For the $n$-th demonstration, we record a time series $\{\mathbf{I}_{n,t}, \mathbf{R}_{n,t}, \boldsymbol{f}_{n,t}, \mathbf{s}_{n,t}\}_t$ at 10 Hz, where $\mathbf{I}_{n,t}$ is the wrist-camera RGB image, $\mathbf{R}_{n,t} {\in} \mathrm{SO}(3)$ is the wrist-camera orientation in the environment frame, $\boldsymbol{f}_{n,t} \in \mathbb{R}^3$ is the net contact force between the robot and environment, and $\mathbf{s}_{n,t}$ encodes task-specific environment state. For example, $s$ is the door opening angle in door opening and the sponge pose relative to the vase in vase wiping.

\vspace{-0mm}
\subsection{Inference of Environment Stiffness}
\vspace{-0mm}
We compute an environment stiffness matrix $\mathbf{K}_e(\mathbf{s}_{n,t})$ for each environment state.
We first normalize each dimension of $\mathbf{s}$ to zero mean and unit variance over the dataset. For each state $\mathbf{s}_{n,t}$, we define a local neighborhood $\mathcal{N}(\mathbf{s}_{n,t})$ as its $k$ nearest neighbors under Euclidean distance in this normalized space. We choose $k$ larger than the task-space dimension so a full stiffness matrix can be estimated. The corresponding contact-force set is $\mathcal{F}(\mathbf{s}_{n,t})
=
\left\{ \boldsymbol{f}_{n',t'} \,\middle|\, \mathbf{s}_{n',t'} \in \mathcal{N}(\mathbf{s}_{n,t}) \right\}$. Note that although $\boldsymbol{f}$ involves forces caused by friction and gravity, because we are using a high-stiffness robot in simulation, contact forces dominate the signal.

Prior work \cite{duan2018learning} infers stiffness from the covariance of the contact forces, $\mathbf{K}_e^2
\propto
\mathbb{E}_{\boldsymbol{f} \in \mathcal{F}(\mathbf{s}_{n,t})}
\bigl[\boldsymbol{f}\,\boldsymbol{f}^\top\bigr]
$. However, this approach depends on how contact directions are sampled: frequently sampled directions contribute more to the covariance and bias the result. Moreover, symmetric positive definite matrices form a curved Riemannian manifold rather than a Euclidean space. Euclidean operations such as element-wise averaging can be geometrically inconsistent and produce the swelling effect, a non-physical increase in determinant that can cause numerical chattering and ``ghost'' stiffness in unobserved directions \cite{arsigny2007geometric}.

Instead, we partition the unit sphere into $m$ fixed angular
sectors (e.g., using a regular grid in spherical coordinates) and aggregate forces within each sector. For the $i$-th sector, we collect forces whose direction falls into that sector. If the sector is non-empty, we define a representative force $\tilde{\boldsymbol{f}}_i$ with the average direction and the 95th percentile magnitude in that sector.
This yields a sector-wise outer product $\tilde{\boldsymbol{f}}_i \tilde{\boldsymbol{f}}_i^\top+\varepsilon \mathbf{I}$, where $\varepsilon \mathbf{I}$ is a small isotropic regularizer that ensures the matrix is full rank.
To combine sector information, we compute a log-Euclidean mean \cite{arsigny2007geometric}. We apply the matrix logarithm to map each matrix to a flat vector space, average in that log domain, and apply the matrix exponential to return to the symmetric positive definite manifold. Concretely,
\begin{align}
\mathbf{S}_e(\mathbf{s}_{n,t})
&=
\frac{1}{m_{\text{valid}}}
\sum_{i=1}^{m_{\text{valid}}}
\log\!\left(
\tilde{\boldsymbol{f}}_i \tilde{\boldsymbol{f}}_i^\top + \varepsilon \mathbf{I}
\right),
\\
\mathbf{K}_e(\mathbf{s}_{n,t})
&\propto
\exp\!\left(
\frac{1}{2}\mathbf{S}_e(\mathbf{s}_{n,t})
\right),
\end{align}
where $m_{\text{valid}}$ is the number of non-empty sectors.

\vspace{-0mm}
\subsection{Computation of Robot Stiffness}
\vspace{-0mm}

After inferring environment stiffness $\mathbf{K}_e(\mathbf{s}_{n,t})$, we construct robot stiffness by making it complementary to the environment: soft where the environment is stiff, and vice versa.

First, we eigendecompose the environment stiffness matrix from the previous section as $\mathbf{K}_e{=} \mathbf{Q}_e \boldsymbol{\Lambda}_e \mathbf{Q}_e^\top$, where $\mathbf{Q}_e {\in} \mathrm{SO}(3)$ gives the principal directions of environment stiffness in the world frame and $\boldsymbol{\Lambda}_e$ contains the corresponding eigenvalues.
To make the robot soft where the environment is stiff, we keep the same eigenvectors and apply a monotonically decreasing mapping $h(\cdot)$ to the eigenvalues. A simple choice is $ h(\lambda_e^j) = 1/(\lambda_e^j + \varepsilon)$, where $\lambda_e^j$ is the diagonal element of $\boldsymbol{\Lambda}_e$ and $\varepsilon$ is a small number. We then normalize the resulting values over the dataset to $[0,1]$. The normalized robot eigenvalues $\tilde{\kappa}^j$ are:
\begin{equation}
    \kappa^j = h(\lambda_e^j), \quad \tilde{\kappa}^j = \mathrm{clamp} \!\left( \frac{\kappa^j - \kappa_{\min}}{\kappa_{\max} - \kappa_{\min}}, \,0,\,1 \right),
\end{equation}
where $\kappa_{\min}$ and $\kappa_{\max}$ are global
lower and upper percentiles of the dataset. We construct the normalized eigenvalue matrix $\tilde{\boldsymbol{\Lambda}}_r = \mathrm{diag}(\tilde{\kappa}^1, \tilde{\kappa}^2, \tilde{\kappa}^3)$.

The policy predicts robot stiffness in the wrist-camera frame. We obtain camera-frame eigenvectors by rotating the world-frame eigenvectors $\mathbf{Q}_e$ by the camera orientation $\mathbf{R}_{n,t}$:
$\mathbf{Q}_r {=} \mathbf{R}_{n,t}^\top \mathbf{Q}_e.$
We then construct the normalized stiffness label 
$\mathbf{\tilde{K}}_r(\mathbf{s}_{n,t}){=} \mathbf{Q}_r \tilde{\boldsymbol{\Lambda}}_r \mathbf{Q}_r^\top.$
By construction, $\mathbf{\tilde{K}}_r$ is symmetric positive definite with eigenvalues in $[0,1]$.


\vspace{-0mm}
\subsection{Training Dataset}
\vspace{-0mm}

This procedure gives training pairs
$\{(\mathbf{I}_{n,t}, \mathbf{\tilde{K}}_r(\mathbf{s}_{n,t}))\}$.
A valid stiffness matrix must be symmetric positive definite. Following prior work \cite{kronander2013learning, abu2018force, kamijo2024learning}, we encode
$\mathbf{\tilde{K}}_r$ using its Cholesky factorization,
$\mathbf{\tilde{K}}_r {=} \mathbf{L}\mathbf{L}^\top$,
where $\mathbf{L}$ is a lower-triangular matrix with strictly positive diagonal entries.

We parametrize the diagonal entries of $\mathbf{L}$ in log-space, so the policy predicts unconstrained diagonal parameters.
The exponential mapping gives a smooth one-to-one correspondence between network outputs and symmetric positive definite stiffness matrices.
\begin{align} \label{eq:policy_output}
\mathbf{L} &=
\begin{bmatrix}
\exp(a_1) & 0          & 0 \\
\ell_{21} & \exp(a_2)  & 0 \\
\ell_{31} & \ell_{32}  & \exp(a_3)
\end{bmatrix},
\end{align}
where $a_1$, $a_2$, $a_3$, $l_{21}$, $l_{31}$, and $l_{32}$ are policy outputs.

\vspace{-0mm}
\subsection{Policy Training}
\vspace{-0mm}

We train a policy that maps a wrist-camera RGB image to a normalized stiffness profile in the camera frame. The input is $\mathbf{I}_{n,t}$, and the output is the unconstrained parameter vector defining the Cholesky factor $\mathbf{L}$ of $\mathbf{\tilde{K}}_r$.

The visual backbone is the small variant of DINOv3 \cite{simeoni2025dinov3}. We use the CLS token as the image feature and freeze the backbone during training. The feature is passed to a feedforward network with 2 fully connected layers of 256 activations. The final layer outputs the 6 scalar values in Eq. \ref{eq:policy_output}.
We train the network with supervised regression on the Cholesky parameters. Given a predicted Cholesky factor $\hat{\mathbf{L}}$ and target factor $\mathbf{L}$, we minimize MSE loss on their parameters. 

To improve generalization and sim-to-real transfer, we apply random cropping, color and brightness jitter, and Gaussian noise during training. We pool all demonstrations and train one policy for the three evaluation tasks.

\vspace{-0mm}
\subsection{Deployment}
\vspace{-0mm}

We run policy inference at 90 Hz. At each step, the policy outputs a normalized stiffness profile $\tilde{\mathbf{K}}_t$. We scale its eigenvalues from $[0,1]$ to $[k_{\min},k_{\max}]$ in N/m to obtain $\mathbf{K}^{\text{raw}}_t$, then smooth $\mathbf{K}^{\text{raw}}_t$ on the symmetric positive definite manifold with a log-domain exponential moving average:
$\mathbf{K}_t{=}\exp\!\left((1-\alpha)\log\!\left(\mathbf{K}_{t-1}+\varepsilon\mathbf{I}\right)+\alpha\log\!\left(\mathbf{K}_t^{\mathrm{raw}}+\varepsilon\mathbf{I}\right)\right)
$, where $\log(\cdot)$ and $\exp(\cdot)$ denote the matrix logarithm and exponential. We set $\alpha=0.2$.

\begin{figure*} [tb]
  \centering
  \includegraphics[width=7.0in]{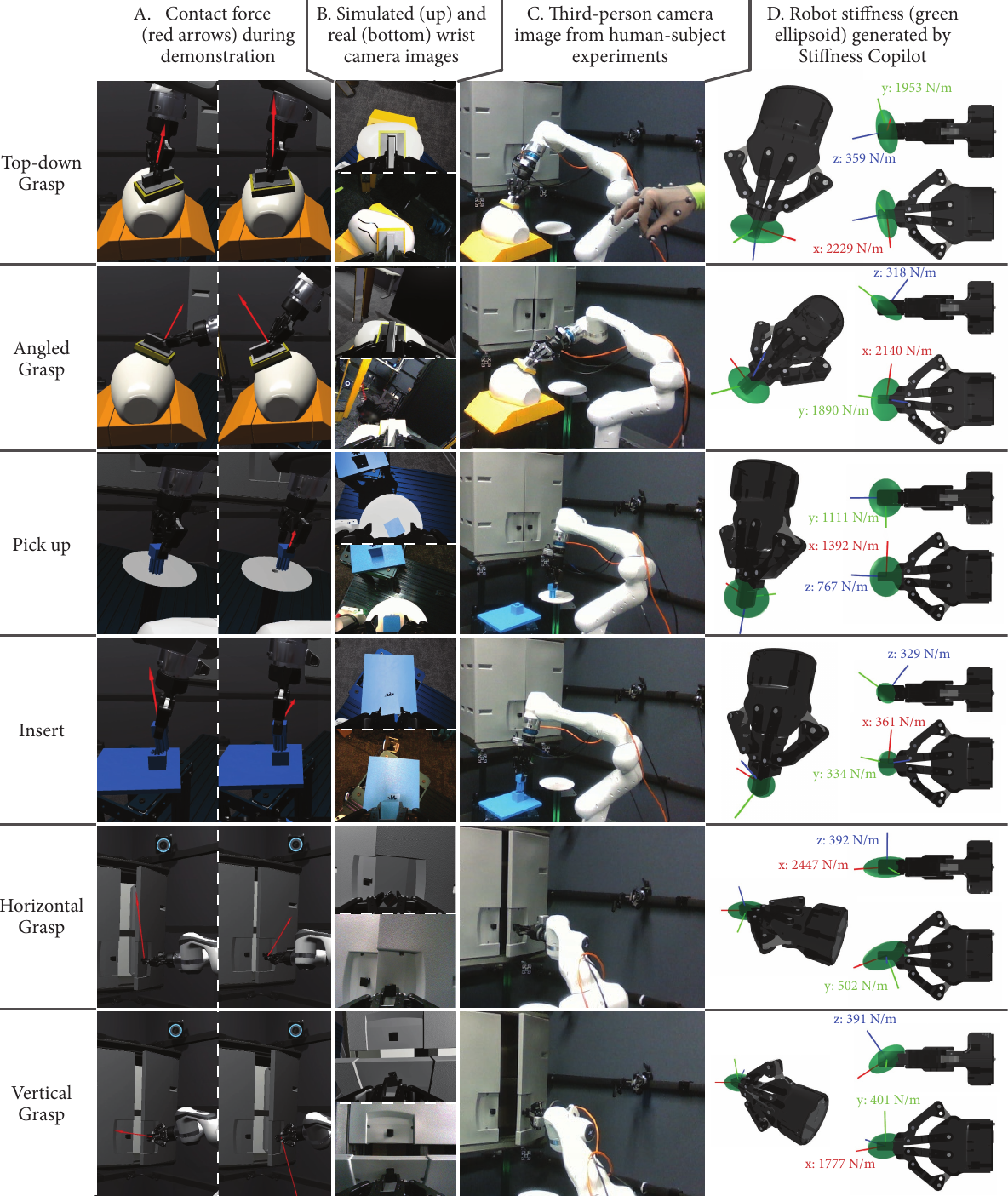}
  \vspace{-3mm}
  \caption{Overview of Stiffness Copilot execution in the human-subject experiment. Each row shows a representative situation.
  (A) We collected contact forces in simulation to infer robot stiffness labels. (B) Stiffness Copilot was trained on simulated wrist-camera images (upper) and deployed on real wrist-camera images (lower), illustrating the sim-to-real gap. (C) Third-person views from our user study, where a participant teleoperated the robot. (D) Stiffness produced by Stiffness Copilot is visualized as a green ellipsoid. A longer ellipsoid axis indicates larger translation stiffness in that direction. We also annotate the stiffness magnitude along each axis. \\
  In Vase Wiping (top two rows), despite different grasp poses, the robot remained compliant perpendicular to the contact surface and stiff in other directions. \\
  In Peg-in-Hole (middle two rows), the robot was stiff while picking up the block and compliant during peg insertion. \\
  In Door Opening (bottom two rows), despite different grasp poses, the robot was stiff along the opening direction and compliant in other directions.}
  \label{fig: ellipsoid}
  \vspace{-2mm}
\end{figure*}

\begin{figure*} [tb]
  \centering
  \includegraphics[width=7.0in]{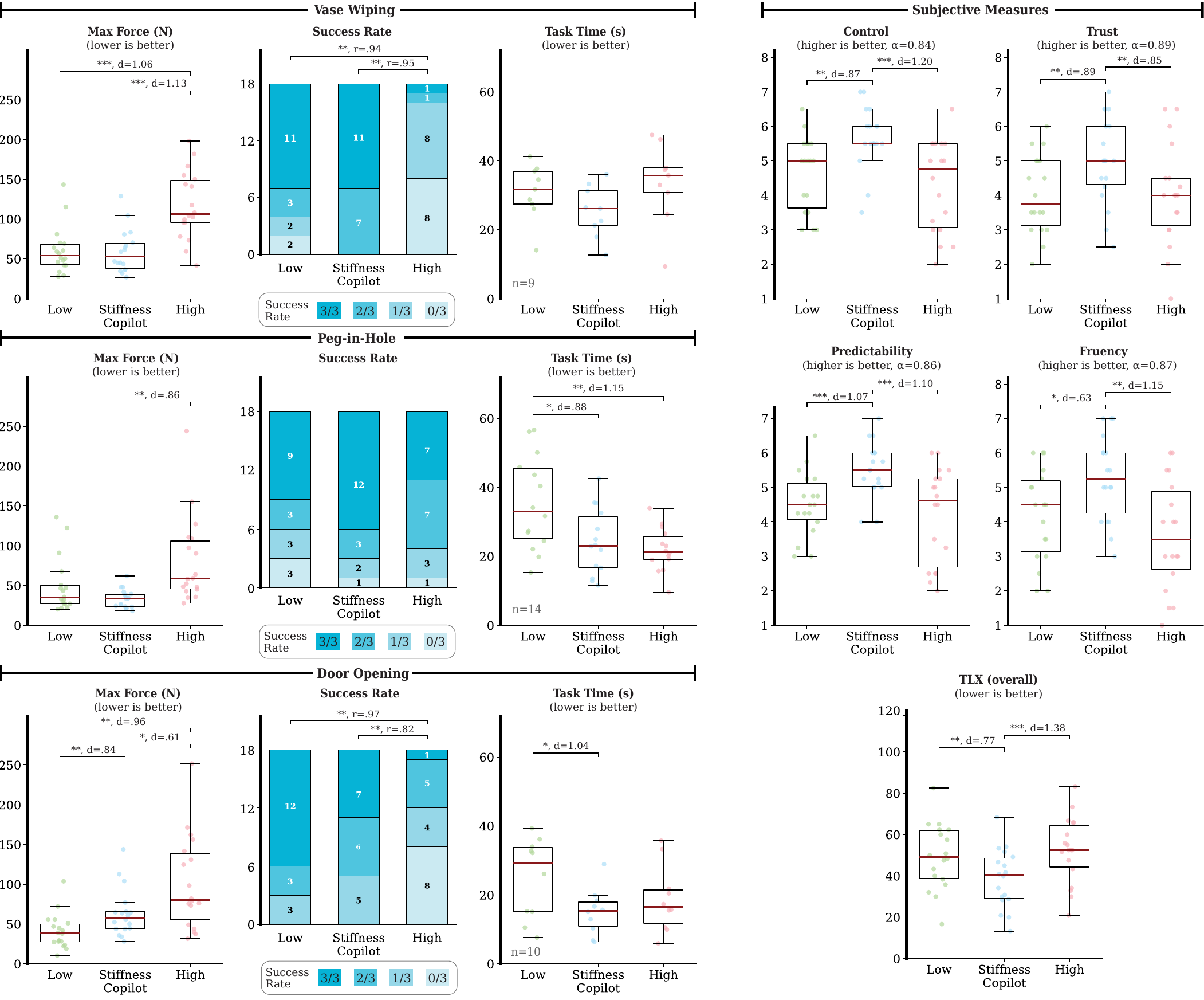}
  \caption{Visualization of our study results. For box-and-whisker plots, each box spans the first to third quartile, the line marks the median, and whiskers extend to at most 1.5 times the interquartile range (IQR). 
  $*/**/{*}{*}{*}$ indicate significance levels \(p<.05/.01/.001\). \(d\) is Cohen's \(d\) for paired \(t\)-tests and \(r\) is the Wilcoxon effect size.
  }
  \vspace{-3mm}
  \label{fig: results}
\end{figure*}

\begin{table*}[ht]
\centering
\caption{Summary of user study results across stiffness conditions. Best values are shown in \textbf{bold}.}
\vspace{-0mm}
\label{tab:study_results}
\begin{tabular}{l|lr|llll}
\toprule
& \multicolumn{2}{c|}{Metric} & Low Stiffness & Stiffness Copilot & High Stiffness & Statistical Test \\
& & & Mean (SD) & Mean (SD) & Mean (SD) & \\
\hline
\multirow{9}{*}{\makecell[l]{Objective\\ Measures}} & \multirow{3}{*}{Vase Wiping} & Max Force (N) & 60.58 (29.23) & \textbf{59.28} (27.04) & 117.46 (42.87) & $F(2,34)=17.15$, $p<.001$$^{***}$, $\eta^2_p=0.502$ \\
& & Success Rate & 0.76 (0.36) & \textbf{0.87} (0.17) & 0.24 (0.28) & $\chi^2=20.13$, $p<.001$$^{***}$, $W=0.559$ \\
& & Time (s) & 30.96 (8.12) & \textbf{25.31} (7.54) & 33.62 (11.57) & $F(2,16)=4.29$, $p<.05$$^{*}$, $\eta^2_p=0.349$ \\
\cline{2-7}
& \multirow{3}{*}{Peg-in-Hole} & Max Force (N) & 49.00 (34.25) & \textbf{33.41} (11.67) & 80.60 (54.44) & $F(2,34)=6.99$, $p<.01$$^{**}$, $\eta^2_p=0.291$ \\
& & Success Rate & 0.67 (0.40) & \textbf{0.81} (0.31) & 0.70 (0.30) & $\chi^2=3.21$, $p=0.201$, $W=0.089$ \\
& & Time (s) & 35.35 (13.59) & 24.22 (9.58) & \textbf{21.95} (6.31) & $F(2,26)=12.01$, $p<.001$$^{***}$, $\eta^2_p=0.480$ \\
\cline{2-7}
& \multirow{3}{*}{\makecell[l]{Door\\ Opening}} & Max Force (N) & \textbf{40.96} (21.97) & 63.63 (29.82) & 101.42 (58.96) & $F(2,34)=11.89$, $p<.001$$^{***}$, $\eta^2_p=0.412$ \\
& & Success Rate & \textbf{0.83} (0.26) & 0.70 (0.28) & 0.31 (0.33) & $\chi^2=19.77$, $p<.001$$^{***}$, $W=0.549$ \\
& & Time (s) & 24.89 (11.68) & \textbf{15.09} (6.67) & 18.63 (9.67) & $F(2,18)=4.99$, $p<.05$$^{*}$, $\eta^2_p=0.357$ \\
\hline
\multirow{5}{*}{\makecell[l]{Subjective\\ Measures}} & \multicolumn{2}{r|}{Control} & 4.64 (1.10) & \textbf{5.67} (0.89) & 4.29 (1.34) & $F(2,34)=10.16$, $p<.001$$^{***}$, $\eta^2_p=0.374$ \\
& \multicolumn{2}{r|}{Trust} & 4.03 (1.14) & \textbf{5.01} (1.27) & 4.01 (1.48) & $F(2,34)=7.22$, $p<.01$$^{**}$, $\eta^2_p=0.298$ \\
& \multicolumn{2}{r|}{Predictability} & 4.51 (0.94) & \textbf{5.51} (0.79) & 4.21 (1.38) & $F(2,34)=11.31$, $p<.001$$^{***}$, $\eta^2_p=0.400$ \\
& \multicolumn{2}{r|}{Fluency} & 4.21 (1.34) & \textbf{5.28} (1.23) & 3.61 (1.59) & $F(2,34)=8.44$, $p<.01$$^{**}$, $\eta^2_p=0.332$ \\
& \multicolumn{2}{r|}{NASA-TLX} & 49.33 (15.89) & \textbf{38.73} (14.15) & 52.43 (16.00) & $F(2,34)=11.36$, $p<.001$$^{***}$, $\eta^2_p=0.401$ \\
\bottomrule
\end{tabular}
\vspace{-7mm}
\end{table*}

\section{Evaluation}  \label{sec:evalution}

Our evaluation tests whether Stiffness Copilot improves task performance and user perception in contact-rich teleoperation. We use constant low and high stiffness as baselines because they span the safety-responsiveness tradeoff and test whether Stiffness Copilot combines the benefits of both ends of the stiffness range.
The project website includes representative videos of each condition and stiffness visualizations.

\vspace{-0mm}
\subsection{Experimental Design \& Conditions}
\vspace{-0mm}

Our user evaluation followed a within-participants design,
with each participant working with the robot under all conditions. Condition order was counterbalanced.

\textit{Low Impedance} (Baseline 1): We configure the robot with a uniformly compliant impedance profile: translational stiffness is fixed at $300$ N/m along all axes, with translational damping set to $20$ N$\cdot$s/m. For orientation control, rotational stiffness is set to $10$ N$\cdot$m/rad and damping to $3.0$ N$\cdot$m$\cdot$s/rad.

\textit{High Impedance} (Baseline 2): The high-impedance baseline maintains a rigid behavior throughout the task, with translational stiffness set to $3000$ N/m in all directions and translational damping increased to $70$ N$\cdot$s/m. Rotational stiffness is set to $100$ N$\cdot$m/rad and damping to $8.0$ N$\cdot$m$\cdot$s/rad.

\textit{Variable Impedance with Stiffness Copilot} (ours): Stiffness Copilot sets the robot translational stiffness. We linearly scale the eigenvalues of the policy output matrix to lie between $300$ and $3000$ N/m. Translational damping is computed in the same eigenspace as $d = 1.4\sqrt{k}$, where $k$ is the corresponding stiffness eigenvalue (N/m) and $d$ is the damping eigenvalue (N$\cdot$s/m). We use a constant medium rotational setting, with stiffness fixed at $40$ N$\cdot$m/rad and damping at $5.5$ N$\cdot$m$\cdot$s/rad.

\vspace{-0mm}
\subsection{Experimental Tasks} \label{sec:tasks}
\vspace{-0mm}

Participants teleoperated the robot to complete 3 contact-rich tasks (Figure \ref{fig: ellipsoid}):

\noindent \textit{Vase Wiping}: Participants grasped a sponge and wiped the marker trace from the vase. This task requires sufficient contact force to remove the trace without triggering the robot's protective stop.

\noindent \textit{Peg-in-Hole}: Participants picked up and inserted a 10-tooth gear-shaped pin into a matching hole with $0.3$ mm clearance. 

\noindent \textit{Cabinet Door Opening}: Participants opened a cabinet door to $90^\circ$ using only the handle. This task requires moving the end effector along the door's opening arc, which is difficult without force feedback; deviations from the arc can produce large interaction forces.

\vspace{-0mm}
\subsection{Implementation Details}
\vspace{-0mm}
We track the operator hand pose with an OptiTrack motion-capture system at 120 Hz. The tracked 6D hand pose is mapped to commands for a Franka Research 3 (FR3) arm using RelaxedIK \cite{rakita2018relaxedik, wang2023rangedik} and executed with the Deoxys operational-space impedance controller \cite{zhu2022viola}. A foot pedal serves as a clutch to enable or pause control.

To compute stiffness labels, one author provided 16 demonstrations per task in simulation. The simulator is built in Unreal Engine with a customized physics solver that uses reduced-order models with data-driven hyper-reduction to simulate contact among rigid and compliant objects at interactive rates \cite{chang2023licrom,zong2023neural,romero2023learning}. This provides rich per-contact state for dexterous manipulation. A Meta Quest 3 was used as the VR display during simulated teleoperation.

In the human-subject experiment, we use a RealSense D405 wrist camera at $640 \times 480$ and 90 Hz. Contact forces are measured with a Bota Systems SensONE Gen-A force-torque sensor sampled at 1000 Hz. Policy rollouts run at 90 Hz on a NVIDIA GeForce 5080 GPU.

\vspace{-0mm}
\subsection{Experimental Procedure}
\vspace{-0mm}

After consent, participants were introduced to the study goal and teleoperation interface. For each condition, they first completed a pick-and-place practice task, then performed three trials of each task in a fixed order: vase wiping, peg-in-hole, and cabinet opening. Each trial had a strict 60 s time limit.
After all nine trials for a condition (3 tasks $\times$ 3 trials), participants completed subjective questionnaires for that condition. They repeated this procedure for all conditions. At the end, participants completed a demographic questionnaire and a semi-structured interview. Participants received 25 USD for approximately 60 minutes.

\vspace{-0mm}
\subsection{Measures} \label{sec:measures}
\vspace{-0mm}
We used objective and subjective measures to evaluate task performance and user experience. For objective measures, we recorded peak contact force from the wrist force-torque sensor in each trial and averaged it across trials. We also calculated task success rates: a trial failed if the emergency stop was triggered or if it exceeded the 60 s time limit. For successful trials, we measured task completion time and report the average across trials. If a participant failed all three trials of a task, we excluded that participant from the completion-time analysis for that task.

For subjective measures, we used questionnaires from prior teleoperation work \cite{wang2023exploiting, wang2024design} to measure perceived control, predictability, fluency, and trust. We used NASA TLX \cite{hart1988development} to measure perceived workload.

\vspace{-0mm}
\subsection{Participants}
\vspace{-0mm}

We recruited 18 participants from a tech company campus (7 females and 11 males), aged 23 to 47 (M${=}31.28$, SD${=}5.44$). Twelve reported prior teleoperation experience. Participants reported moderate-to-high familiarity with robots (M${=}4.64$, SD${=}2.04$) and 3D video games (M ${=}5.17$, SD${=}2.09$) on a 7-point Likert scale. Most participants were right-handed (15 right-handed, 3 left-handed), and all operated the robot with their right hand.

\vspace{-0mm}
\subsection{Results}
\vspace{-0mm}

Figure \ref{fig: results} and Table \ref{tab:study_results} summarize our results. We analyzed continuous measures with a one-way repeated-measures ANOVA, followed by Holm-corrected paired t-tests. We analyzed success rate with a Friedman test followed by Holm-corrected Wilcoxon signed-rank tests.

\textit{Task Performance} -- Stiffness Copilot led to smaller contact forces than the high-stiffness condition across all three tasks (all $p{<}.05$ and Cohen's $d {>}.6$). It also achieved higher success rates than high stiffness across tasks, with significant differences for Vase Wiping and Door Opening ($p{<}.01$ for both), but not for Peg-in-Hole insertion. One possible explanation is that peg-in-hole insertion is more perception-limited and requires precise visual alignment before insertion. This bottleneck may dominate performance and reduce the observable impact of robot impedance. Compared to low stiffness, Stiffness Copilot reduced task time across all tasks, with significant reductions for Peg-in-Hole and Cabinet Opening (both $p{<}.05$ and $d{>}.8$), but not for Vase Wiping ($p{=}.091$ and $d{=}.72$). We attribute this non-significant result to reduced statistical power, as only $n{=}9$ participants were included in the time analysis after excluding participants who failed all three trials (see \cref{sec:measures}).

\textit{User Perception} -- Our results revealed that participants perceived the robot with Stiffness Copilot as
significantly more under control, predictable, fluent, and
trustworthy, and reported significantly lower workload (all
$p{<}.05$) with medium-to-large effect sizes (all $d{>}.6$), compared to both the high stiffness and low stiffness robot.

In the post-experiment interview, participants often described the low-stiffness robot as slow, flowy, or delayed.
They also described the high-stiffness condition as difficult. For example,
P4 commented that ``\textit{On the [high-stiffness] condition, I was trying a lot harder to try to follow the path of the cabinet door better to open it up. For the [Stiffness Copilot] condition, I didn't need to think about this.}''
P1 commented that 
``\textit{[Stiffness Copilot] was as responsive as the [High Stiffness] condition, but it felt more in control. 
}''
%



\textit{Summary} -- Stiffness Copilot combines the benefits of both baselines. It provides collision safety comparable to low stiffness, reflected by lower contact forces, while allowing operators to complete tasks as quickly as in high stiffness. Participants also rated Stiffness Copilot as significantly more controllable, trustworthy, predictable, and fluent than either baseline.

\vspace{-0mm}
\section{Discussion}
\vspace{-0mm}

We proposed Stiffness Copilot, a vision-based policy that takes wrist-camera images as input and outputs a direction-dependent translational stiffness matrix. It enables a shared-control teleoperation paradigm: the operator commands target poses while the policy modulates robot stiffness online. In a user study, Stiffness Copilot matched the collision safety of a low-stiffness baseline and the task efficiency of a high-stiffness baseline, while improving user experience.

Several limitations motivate future work. 
First, we only modulate translational impedance and keep rotational impedance fixed; extending the policy to adapt rotational stiffness and damping is an important next step. 

Second, the current policy uses only wrist-camera images. This can miss contact constraints that are occluded or cannot be inferred from appearance. Future work could combine visual stiffness prediction with force or tactile feedback to refine impedance online after contact, as in recent tactile-reactive and contact-grounded policies \cite{xu2024letacmpc,xu2026contactgrounded}.
Third, our current supervision and training data focus on quasi-static contact interactions. Future work should extend Stiffness Copilot to more dynamic regimes, such as hammering, where high stiffness supports fast and accurate swings, while lower impact-time stiffness reduces force transmission to the arm. 
Fourth, although we evaluate Stiffness Copilot in a shared-control setting, the same mechanism could be integrated into autonomous manipulation policies to adapt impedance during execution. 
Finally, while we use a single policy across three tasks, we do not evaluate generalization to unseen objects or tasks, such as opening an unseen door. Scaling to broader settings may require more diverse training data and more expressive architectures so one policy can adjust impedance across a wider range of contact-rich tasks.

\bibliography{root}
\bibliographystyle{IEEEtran}

\end{document}